\documentclass{article}

\usepackage{arxiv}
\usepackage[round]{natbib}
\usepackage[colorlinks=true,allcolors=blue]{hyperref}
\usepackage[utf8]{inputenc} %
\usepackage[T1]{fontenc}    %
\usepackage{url}            %
\usepackage{booktabs}       %
\usepackage{amsfonts}       %
\usepackage{nicefrac}       %
\usepackage{microtype}      %

\usepackage{graphicx}
\usepackage{subfig}
\usepackage{color}
\usepackage{multirow}
\usepackage{array}
\usepackage{amsopn}
\usepackage{wrapfig}
\usepackage{amsmath}
\usepackage{setspace}

\def\R{\mathbb{R}}

\DeclareMathOperator*{\logadd}{logadd}

\title{Letter-Based Speech Recognition\\
with Gated ConvNets}

\author{
  Vitaliy Liptchinsky\\
  Facebook AI Research\\
  \texttt{vitaliy888@fb.com} \\
  \And
  Gabriel Synnaeve\\
  Facebook AI Research\\
  \texttt{gab@fb.com}\\
  \And
  Ronan Collobert\\
  Facebook AI Research\\
  \texttt{locronan@fb.com}\\
}

\begin{document}

\maketitle

\begin{abstract}
  In the recent literature, ``end-to-end'' speech systems often refer to
  letter-based acoustic models trained in a sequence-to-sequence manner,
  either via a recurrent model or via a structured output learning approach
  (such as CTC~\citep{graves2006connectionist}). In contrast to traditional
  phone (or senone)-based approaches, these ``end-to-end'' approaches
  alleviate the need of word pronunciation modeling, and do not require a
  ``forced alignment'' step at training time. Phone-based approaches remain
  however state of the art on classical benchmarks. In this paper, we
  propose a letter-based speech recognition system, leveraging a ConvNet
  acoustic model. Key ingredients of the ConvNet are Gated Linear Units and
  high dropout. The ConvNet is trained to map audio sequences to their
  corresponding letter transcriptions, either via a classical CTC approach,
  or via a recent variant called ASG~\citep{collobert2016wav2letter}. Coupled
  with a simple decoder at inference time, our system matches the best existing
  letter-based systems on WSJ (in word error rate), and shows near
  state of the art performance on LibriSpeech~\citep{panayotov2015librispeech}.

\end{abstract}

\section{Introduction}

State of the art speech recognition systems leverage pronunciation
models as well as speaker adaptation techniques involving speaker-specific
features. These systems rely on lexicon dictionaries, which decompose
words into one or more sequences of phones. Phones themselves are
decomposed into smaller sub-word units, called senones. Senones are
carefully selected through a procedure involving a phonetic-context-based
decision tree built from another GMM/HMM system. In the recent literature,
``end-to-end'' speech systems attempt to break away from these hardcoded
a-priori, the underlying assumption being that with enough data
pronunciations should be implicitly inferred by the model, and speaker
robustness should be also achieved. A number of works have thus naturally
proposed ways how to learn to map audio sequences directly to their
corresponding letter sequences. Recurrent models,
structured-output learning or combination of both are the main contenders.

In this paper, we show that simple convolutional neural networks (CNNs)
coupled with structured-output learning can outperform existing letter-based
solutions. Our CNNs employ Gated Linear Units (GLU). Gated ConvNets have
been shown to reduce the vanishing gradient problem, as they provide a
linear path for the gradients while retaining non-linear capabilities,
leading to state of the art performance both in natural language modeling
and machine translation tasks~\citep{dauphin2017lm,gehring2017mt}. We train
our system with a structured-output learning approach, either with
CTC~\citep{graves2006connectionist} or ASG~\citep{collobert2016wav2letter}.
Coupled with a custom-made simple beam-search decoder, we exhibit word
 error rate (WER) performance matching the best existing letter-based
 systems, both for the WSJ and LibriSpeech datasets~\citep{panayotov2015librispeech}. 
 While phone-based systems still
lead on WSJ ($81$h of labeled data), our system is competitive with the
existing state of the art systems on LibriSpeech ($960$h).

The rest of the paper is structured as follows: the next section goes over
the history of the work in the automatic speech recognition area.
We then detail the convolutional
networks used for acoustic modeling, along with the structured-output
learning and decoding approaches. The last section shows experimental
results on WSJ and LibriSpeech.

\begin{figure*}
  \vspace*{-0.3cm}
  \centering
  \includegraphics[width=0.8\linewidth]{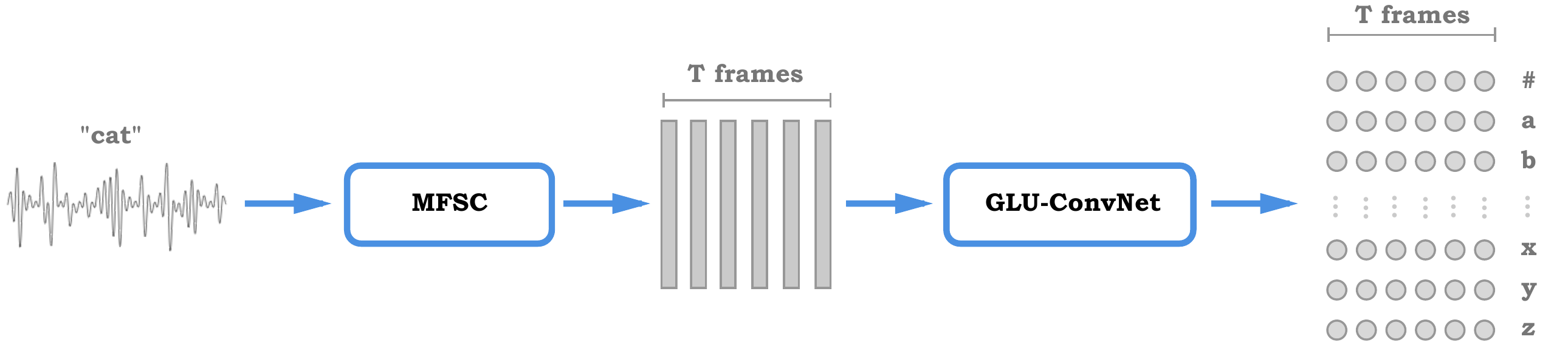}
  \caption{
    \label{fig-overview}
    Overview of our acoustic model, which computes log-mel filterbanks (MFSC) that are
    fed to a Gated ConvNet. The ConvNet outputs one score for each letter in
    the dictionary, and for each input feature frame. At inference time, these
    scores are fed to a decoder (see Section~\ref{sec-decoder}) to form
    the most likely sequence of words. At training time, the scores are fed
    to the CTC or ASG criterions (see Figure~\ref{fig-asg}) which promote
    sequences of letters leading to the target transcription sequence (here ``c a
    t'').
  }
\end{figure*}

\section{Background}

The historic pipeline for speech recognition requires first
training an HMM/GMM model to force align the units on which the final
acoustic model operates (most often context-dependent phone or senone
states) \cite{woodland1993htk}. The performance improvements brought
by deep neural networks (DNNs)
\citep{mohamed2012acoustic,hinton2012deep} and convolutional neural
networks (CNNs) \citep{sercu2015very,soltau2014joint} for acoustic
modeling only extend this training pipeline.  Current state
of the art models on LibriSpeech also employ this approach
\citep{panayotov2015librispeech,peddinti2015time}, with an additional
step of speaker adaptation \citep{saon2013speaker,peddinti2015jhu}.
Departing from this historic pipeline,
\cite{senior2014gmm} proposed GMM-free training, but the approach
still requires to generate a forced alignment. Recently, maximum
mutual information (MMI) estimation \citep{bahl1986icassp} was used to
train neural network acoustic models~\citep{povey2016mmi}.
The MMI criterion~\citep{bahl1986icassp}
maximizes the mutual information between the acoustic sequence and
word sequences or the Minimum Bayes Risk (MBR)
criterion~\citep{gibson2006interspeech}, and belongs to segmental
discriminative training criterions, although compatible with
generative models.

Even though connectionist approaches~\citep{lee1995rnn,lecun1995convolutional} 
long coexisted with HMM-based approaches, they had a recent
resurgence. A modern work that
directly cut ties with the HMM/GMM pipeline used a recurrent
neural network (RNN) \citep{graves2013speech} for phoneme transcription
with the connectionist temporal classification (CTC) sequence loss
\citep{graves2006connectionist}. This approach was then extended
to character-based systems \citep{graves2014towards} and improved with
attention
mechanisms~\citep{bahdanau2016attention,chan2016listenattendspell}. But
the best such systems are often still behind state of the art
phone-based (or senone-based) systems. Competitive end-to-end
approaches leverage acoustic models (often ConvNet-based) topped with
RNN layers as in
\citep{hannun2014deep,miao2015eesen,saon2015ibm,amodei2015deep,zhou2018policy,zeyer2018improved}
(e.g. a state of the art on WSJ \citep{chan2015deep}), trained with a
sequence criterion (the most popular ones being
CTC~\citep{graves2006connectionist} and MMI~\citep{bahl1986icassp}).
A survey of segmental models
can be found in~\citep{tang2017segmental}.  On conversational speech
(that is not the topic of this paper), the state of the art is still
held by complex ConvNets+RNNs acoustic models (which are also trained
or refined with a sequence criterion), coupled with domain-adapted
language models
\citep{xiong2017microsoft,saon2017english}.

\section{Architecture}

Our acoustic model (see an overview in Figure~\ref{fig-overview}) is a
Convolutional Neural Network (ConvNet) \citep{lecun1995convolutional}, with
Gated Linear Units (GLUs)~\citep{dauphin2017lm} and dropout applied to activations
of each layer except the output one. The model is fed with
log-mel filterbank features, and is trained with either the Connectionist
Temporal Classification (CTC) criterion \citep{graves2006connectionist}, or with the ASG criterion: a variant of CTC that does not have blank labels but employs a simple duration model through letter transition scores~\citep{collobert2016wav2letter}. At inference, the acoustic model is coupled
with a decoder which performs a beam search, constrained with a count-based
language model. We detail each of these components in the following.

\subsection{Gated ConvNets for Acoustic Modeling}
\label{sec-glu}

The acoustic model architecture is a 1D Gated Convolutional Neural Network
(Gated ConvNet), trained to map a sequence of audio features to its
corresponding letter transcription. Given a dictionary of letters ${\cal
  L}$, the ConvNet (which acts as a sliding-approach over the input
sequence) outputs one score for each letter in the dictionary, for each
input frame. In the transcription, words are separated by a special letter,
denoted \texttt{\small \#}.

1D ConvNets were introduced early in the speech community, and are also
referred as Time-Delay Neural Networks
(TDNNs)~\citep{waibel1989tdnn}. Gated ConvNets~\citep{dauphin2017lm} stack
1D convolutions with Gated Linear Units. More formally, given an input
sequence $\mathbf{X} \in \R^{T\times d^i}$ with $T$ frames of
$d$-dimensional vectors, the $i^{\rm th}$ layer of our network performs the
following computation:
\begin{equation}
\label{eq-glu}
h^i(\mathbf{X}) = (\mathbf{X}\ast\mathbf{W}^i+\mathbf{b}^i)\otimes \sigma(\mathbf{X}\ast\mathbf{V}^i+\mathbf{c}^i)\,,
\end{equation}
where $\ast$ is the convolution operator, $\mathbf{W}^i,\, \mathbf{V}^i \in \R^{d^{i+1}\times d^i \times k^i}$ and $\mathbf{b}^i,\, \mathbf{c}^i \in \R^{d^{i+1}}$ are the
learned parameters (with convolution kernel size $k^i$), $\sigma(\cdot)$ is the sigmoid function and $\otimes$ is the element-wise product between matrices.

Gated ConvNets have been shown to reduce the vanishing gradient problem, as
they provide a linear path for the gradients while retaining non-linear
capabilities, leading to state of the art performance both for natural
language modeling and machine translation
tasks~\citep{dauphin2017lm,gehring2017mt}.

\subsubsection{Feature Normalization and Zero-Padding}

Each input feature sequence is normalized to have zero mean and unit variance.
Given an input sequence $\mathbf{X} \in \R^{T\times d}$, a convolution
with kernel size $k$ will output $T-k+1$ frames, due to border effects. To
compensate those border effects, we pad the log-mel filterbanks $\mathbf{X}^0$
with zeroed frames. To take into account the whole network, the padding size
is $\sum_i (k^i-1)$, divided into two equal parts at the beginning and the
end of the sequence.

\subsection{Acoustic Model Training}

\begin{figure*}[h]
  \centering
  \subfloat[]{\label{fig-ctc-ug}\includegraphics[width=0.3\linewidth]{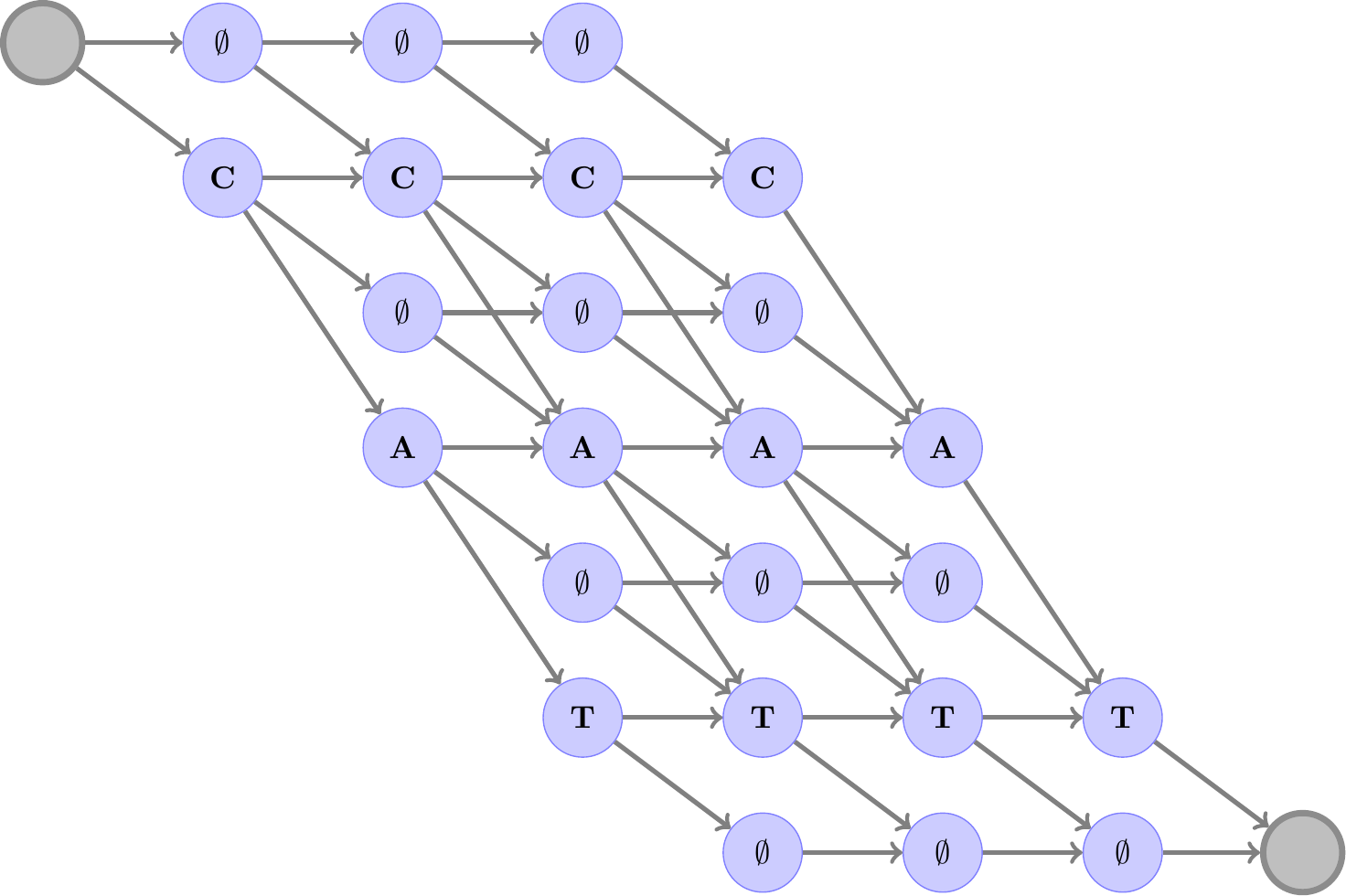}}
  \hspace*{0.1cm}
  \subfloat[]{\label{fig-asg-ug}\includegraphics[width=0.3\linewidth]{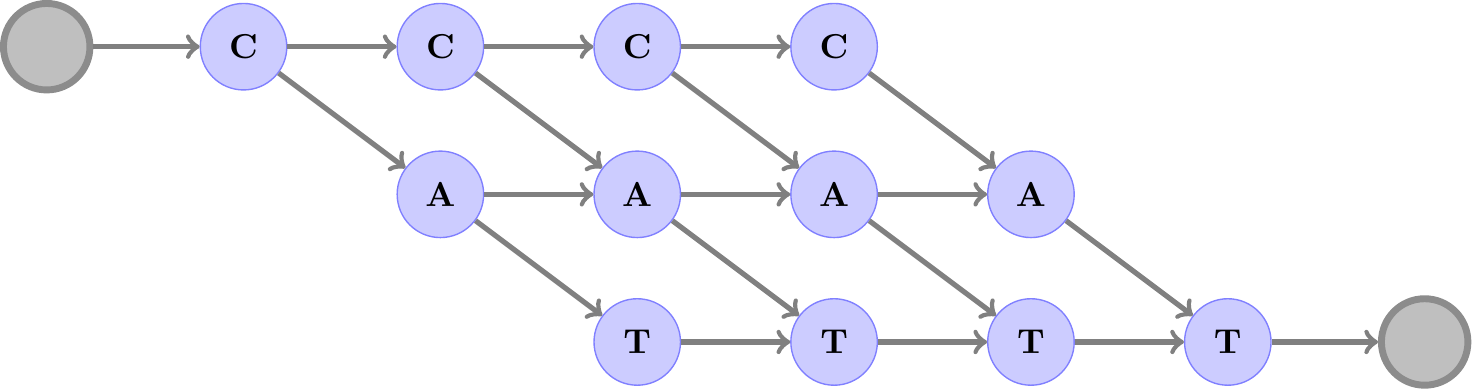}}
  \hspace*{0.1cm}
  \subfloat[]{\label{fig-full-ug}\includegraphics[width=0.3\linewidth]{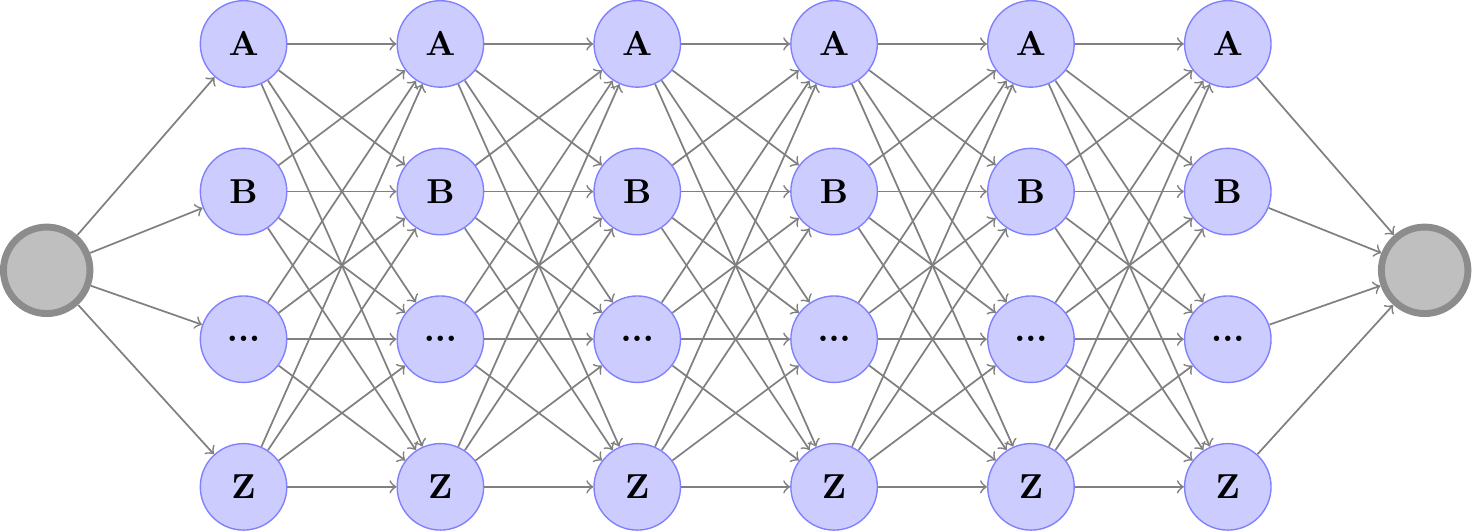}}
  \caption{
    \label{fig-asg}
    (a) The CTC graph which represents all the acceptable sequences of
    letters for the transcription ``cat'' over 6 frames. (b) The same graph
    used by ASG, where blank labels have been discarded. (c) The fully
    connected graph describing all possible sequences of letter; this graph
    is used for normalization purposes in ASG. Un-normalized transitions
    scores are possible on edges of these graphs. At each time step, nodes
    are assigned a conditional un-normalized score, output by the Gated
    ConvNet acoustic model.}
\end{figure*}
We considered two structured-output learning approaches to train our
acoustic models: the Connectionist Temporal Classification (CTC), and a variant
called AutoSeG (ASG).

\subsubsection{The CTC Criterion}
\label{sec-ctccriterion}

CTC~\citep{graves2006connectionist} efficiently enumerates all possible
sequences of sub-word units (e.g. letters) which can lead to the correct
transcription, and promotes the score of these sequences. CTC also allows a
special ``blank'' state to be optionally inserted between each sub-word
unit. The rationale behind the blank state is two-fold: (i) modeling
``garbage'' frames which might occur between each letter and (ii)
identifying the separation between two identical consecutive sub-word units
in a transcription. Figure~\ref{fig-ctc-ug} shows the CTC graph describing
all the possible sequences of letters leading to the word ``cat'', over 6
frames. We denote ${\cal G}_{ctc}(\theta, T)$ the CTC acceptance graph over
$T$ frames for a given transcription $\theta$, and $\pi = {\pi_1,\,\dots,\,
  \pi_T} \in {\cal G}_{ctc}(\theta, T)$ a path in this graph representing a
(valid) sequence of letters for this transcription. CTC assumes that the
network outputs probability scores, normalized at the frame level. At each
time step $t$, each node of the graph is assigned with its corresponding
log-probability letter $i$ (that we denote $f^t_i(\mathbf{X})$) output by
the acoustic model (given an acoustic sequence $\mathbf{X}$). CTC minimizes
the Forward score over the graph ${\cal G}_{ctc}(\theta, T)$:
\begin{equation}
  \label{eq-ctc}
  CTC(\theta, T) = - \logadd_{\pi \in {\cal G}_{ctc}(\theta, T)} \sum_{t=1}^T f^t_{\pi_t}(\mathbf{X})\,,
\end{equation}
where the ``logadd'' operation (also called ``log-sum-exp'') is
defined as $\logadd(a, b) = \log(\exp(a) + \exp(b))$. This overall score
can be efficiently computed with the Forward algorithm.

\subsubsection{The ASG Criterion}
\label{sec-autosegcriterion}

Blank labels introduce code complexity when decoding letters into words. Indeed,
with blank labels ``\o'', a word gets many entries in the sub-word unit
transcription dictionary~(e.g. the word ``cat'' can be represented as ``c a
t'', ``c \o\ a t'', ``c \o\ a t'', ``c \o\ a \o\ t'', etc... -- instead of
only ``c a t''). We replace the blank label by special letters modeling
repetitions of preceding letters. For example ``caterpillar'' can be
written as ``caterpil1ar'', where ``1'' is a label to represent one
repetition of the previous letter.

The AutoSeG (ASG) criterion~\citep{collobert2016wav2letter} removes the
blank labels from the CTC acceptance graph ${\cal G}_{ctc}(\theta, T)$
(shown in Figure~\ref{fig-ctc-ug}) leading to a simpler graph that we
denote ${\cal G}_{asg}(\theta, T)$ (shown in Figure~\ref{fig-asg-ug}). In
contrast to CTC which assumes per-frame normalization for the acoustic
model scores, ASG implements a sequence-level normalization to prevent the
model from diverging (the corresponding graph enumerating all possible
sequences of letters is denoted ${\cal G}_{asg}(\theta, T)$, as shown in
Figure~\ref{fig-full-ug}). ASG also uses unnormalized transition scores
$g_{i,j}(\cdot)$ on each edge of the graph, when moving from label $i$ to
label $j$, that are trained jointly with the acoustic model. 
This leads to the following criterion::
\begin{equation}
  \label{eq-asg}
  \begin{split}
    ASG(\theta, T) = & - \logadd_{\pi \in {\cal G}_{asg}(\theta, T)} \sum_{t=1}^T (f^t_{\pi_t}(\mathbf{X})+g_{\pi_{t-1},\pi_t}(\mathbf{X})) \\
    & + \logadd_{\pi \in {\cal G}_{full}(\theta, T)} \sum_{t=1}^T (f^t_{\pi_t}(\mathbf{X})+g_{\pi_{t-1},\pi_t}(\mathbf{X})) \,.
  \end{split}
\end{equation}
The left-hand part in Equation~$(\ref{eq-asg})$ promotes the score of
letter sequences leading to the right transcription (as in
Equation~(\ref{eq-ctc}) for CTC), and the right-hand part demotes the score
of all sequences of letters.
As for CTC, these two parts can be efficiently computed with the Forward
algorithm.

When removing transitions in Equation~(\ref{eq-asg}), the sequence-level
normalization becomes equivalent to the frame-level normalization found in
CTC, and the ASG criterion is mathematically equivalent to CTC with no
blank labels. However, in practice, we observed that acoustic models
trained with a transition-free ASG criterion had a hard time to converge.

\subsubsection{Other Training Considerations}
\label{sec-dropout}

We apply dropout at the output to all layers of the acoustic model. Dropout
retains each output with a probability $p$, by applying a multiplication
with a Bernoulli random variable taking value $1$ with probability $p$
and $0$ otherwise~\citep{srivastava2014dropout}.

Following the original implementation of Gated
ConvNets~\citep{dauphin2017lm}, we found that using both weight
normalization~\citep{salimans2016wn} and gradient
clipping~\citep{pascanu2013rnn} were speeding up training convergence. The clipping
we implemented performs:
\begin{equation}
  \label{eq-clipping}
\overset{\sim}{\nabla C} =  \max(||\nabla C||, \epsilon) \frac{\nabla C}{||\nabla C||}\,,
\end{equation}
where $C$ is either the CTC or ASG criterion, and $\epsilon$ is some
hyper-parameter which controls the maximum amplitude of the gradients.

\subsection{Beam-Search Decoder}
\label{sec-decoder}
We wrote our own one-pass decoder, which performs a simple beam-search with
beam thresholding, histogram pruning and language model
smearing~\citep{steinbiss1994improvements}. We kept the decoder as simple as
possible (under 1000 lines of C code). We did not implement any sort of
model adaptation before decoding, nor any word graph rescoring. Our decoder
relies on KenLM~\citep{heafield2013scalable} for the language modeling
part. It also accepts unnormalized acoustic scores (transitions and
emissions from the acoustic model) as input. The decoder attempts to maximize
the following:
\begin{equation}
  \label{eq-decoder}
  \begin{split}
    {\cal L}(\theta) =  & \logadd_{\pi \in {\cal G}_{lex}(\theta, T)} \sum_{t=1}^T (f_{\pi_t}(x)+g_{\pi_{t-1},\pi_t}(x)) \\
                        & + \alpha \log P_{lm}(\theta) + \gamma |\{i| \pi_i = \textrm{\texttt{\small \#}}\}|\,,
    \end{split}
\end{equation}
where ${\cal G}_{lex}(\theta, T)$ is a graph constrained by lexicon,
$P_{lm}(\theta)$ is the probability of the language model given a
transcription $\theta$, $\alpha$, $\beta$, and $\gamma$ are three
hyper-parameters which control the weight of the language model, 
and the silence (\#) insertion penalty, respectively.

The beam of the decoder tracks paths with highest scores according to
Equation~(\ref{eq-decoder}), by bookkeeping pairs of (language model,
lexicon) states, as it goes through time. The language model state
corresponds to the $(n-1)$-gram history of the $n$-gram language model,
while the lexicon state is the sub-word unit position in the current word
hypothesis. To maintain diversity in the beam, paths with identical
(language model, lexicon) states are merged. Note that traditional decoders
combine the scores of the merged paths with a $\max(\cdot)$ operation (as
in a Viterbi beam-search) -- which would correspond to a $\max(\cdot)$
operation in Equation~(\ref{eq-decoder}) instead of $\logadd(\cdot)$.  We
consider instead the $\logadd(\cdot)$ operation (as first suggested
by~\cite{bottou1991phd}), as it takes into account the contribution of all
the paths leading to the same transcription, in the same way we do during
training (see Equation~(\ref{eq-asg})). In
Section~\ref{sec-experiments-arch}, we show that this leads to better
accuracy in practice.

\section{Experiments}

We benchmarked our system on WSJ (about $81$h of labeled audio data) and
LibriSpeech~\citep{panayotov2015librispeech} (about $960$h). We kept the
original 16 kHz sampling rate. For WSJ, we considered the classical setup
\textsc{si284} for training, \textsc{dev93} for validation, and
\textsc{eval92} for evaluation.  For LibriSpeech, we considered the two
available setups \textsc{clean} and \textsc{other}. All the
hyper-parameters of our system were tuned on validation sets. Test sets
were used only for the final evaluations.

The letter vocabulary ${\cal L}$ contains 30 graphemes: the standard
English alphabet plus the apostrophe, silence (\textsc{\small \#}), and two
special ``repetition'' graphemes which encode the duplication (once or
twice) of the previous letter (see
Section~\ref{sec-autosegcriterion}). Decoding is achieved with our own
decoder (see Section~\ref{sec-decoder}). We used standard language models
for both datasets, \emph{i.e.} a 4-gram model (with about $165K$ words in
the dictionary) trained on the provided data for WSJ, and a 4-gram
model\footnote{\url{http://www.openslr.org/11}.}  (about $200K$ words) for
LibriSpeech. In the following, we either report letter-error-rates (LERs)
or word-error-rates (WERs).

Training was performed with stochastic gradient descent on WSJ, and
mini-batches of $4$ utterances on LibriSpeech. Clipping parameter (see
Equation~(\ref{eq-clipping})) was set to $\epsilon=0.2$. We used a momentum
of $0.9$. Input features, log-mel filterbanks, were computed with 40 coefficients,
a 25 ms sliding window and 10 ms stride.

We implemented everything using \textsc{\small
  Torch7}\footnote{\url{http://www.torch.ch}.}.
The CTC and ASG criterions, as well as the decoder were implemented in C
(and then interfaced into \textsc{\small Torch}).

\begin{table*}
\begin{center}
  \caption{
    Architecture details. ``\#conv.'' is the number of convolutional
    layers. Dropout amplitude, ``\#hu'' (number of output hidden units) and
    ``kw'' (convolution kernel width) are provided for the first and last
    layer (all are linearly increased with layer depth). The size of the final
    layer is also provided.}
    \label{tbl-arch}
      \small
      \setlength{\tabcolsep}{4.0pt}
      \begin{tabular}{lccccccccc}
        \toprule
        Dataset & Architecture & \#conv. & dropout & \#hu & kw & \#hu \\
        & & & {\footnotesize first/last layer} & {\footnotesize first/last layer} & {\footnotesize first/last layer} & {\footnotesize full connect} \\
        \midrule
        WSJ & Low Dropout  & 17 & 0.25/0.25 & 100/375  & 3/21 & 1000 \\
        \midrule
        LibriSpeech & Low Dropout  & 17 & 0.25/0.25 & 200/750  & 13/27 & 1500 \\
                    & High Dropout & 19 & 0.20/0.60 & 200/1000 & 13/29 & 2000 \\
        \bottomrule
      \end{tabular}
\end{center}
\end{table*}

\subsection{Architecture}
\label{sec-experiments-arch}

We tuned our acoustic model architectures by grid search, validating on the
dev sets. We consider here two architectures, with low and high amount of
dropout (see the parameter $p$ in Section~\ref{sec-dropout}).
Table~\ref{tbl-arch} reports the details of our architectures. The amount
of dropout, number of hidden units, as well as the convolution kernel width
are increased linearly with the depth of the neural network. Note that as
we use Gated Linear Units (see Section~\ref{sec-glu}), each layer is
duplicated as stated in Equation~(\ref{eq-glu}). Convolutions are followed
by a fully connected layer, before the final layer which outputs $30$
scores (one for each letter in the dictionary). Concerning WSJ, the
\textsc{Low Dropout} ($p=0.2$) architecture has about $17$M trainable
parameters. For LibriSpeech, architectures have about $130$M and
$208$M of parameters for the \textsc{Low Dropout} ($p=0.2$) and
\textsc{High Dropout} ($p=0.2\rightarrow 0.6$) architectures, respectively.

\begin{figure}
  \centering
  \subfloat[]{\label{fig-ler-wer-clean}\includegraphics[width=0.49\linewidth]{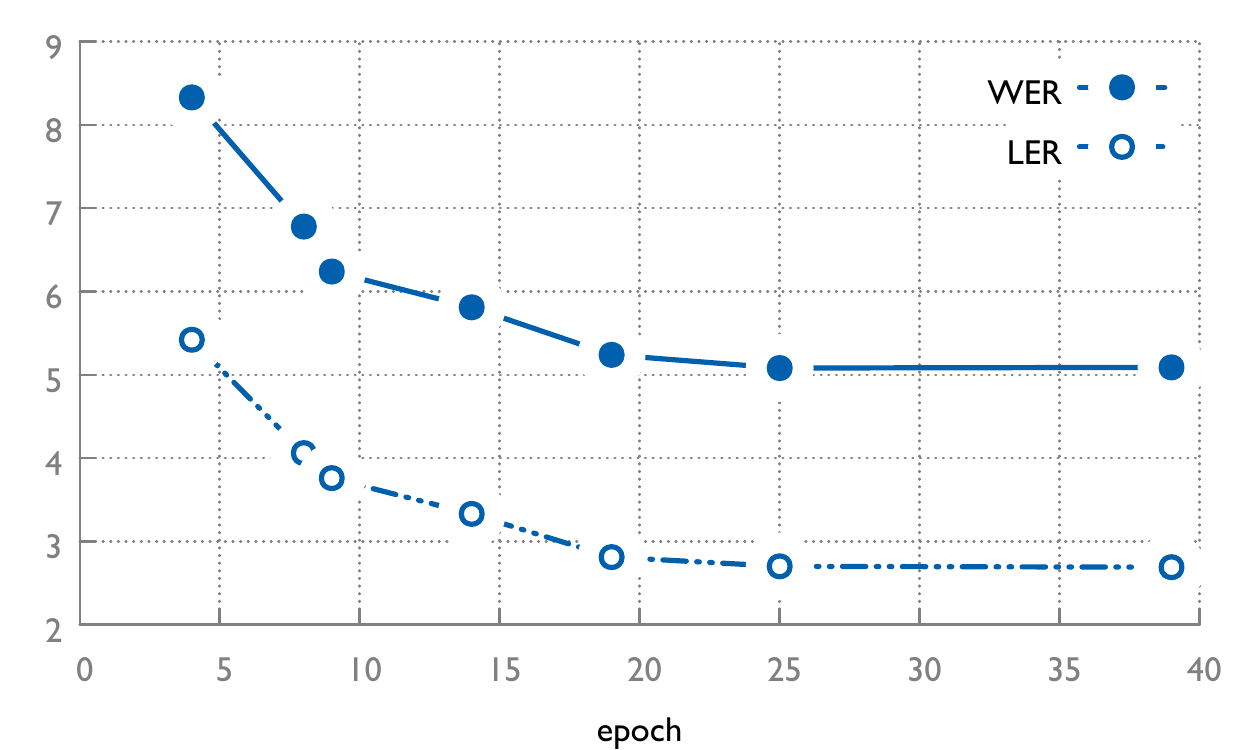}}
  \subfloat[]{\label{fig-ler-wer-other}\includegraphics[width=0.49\linewidth]{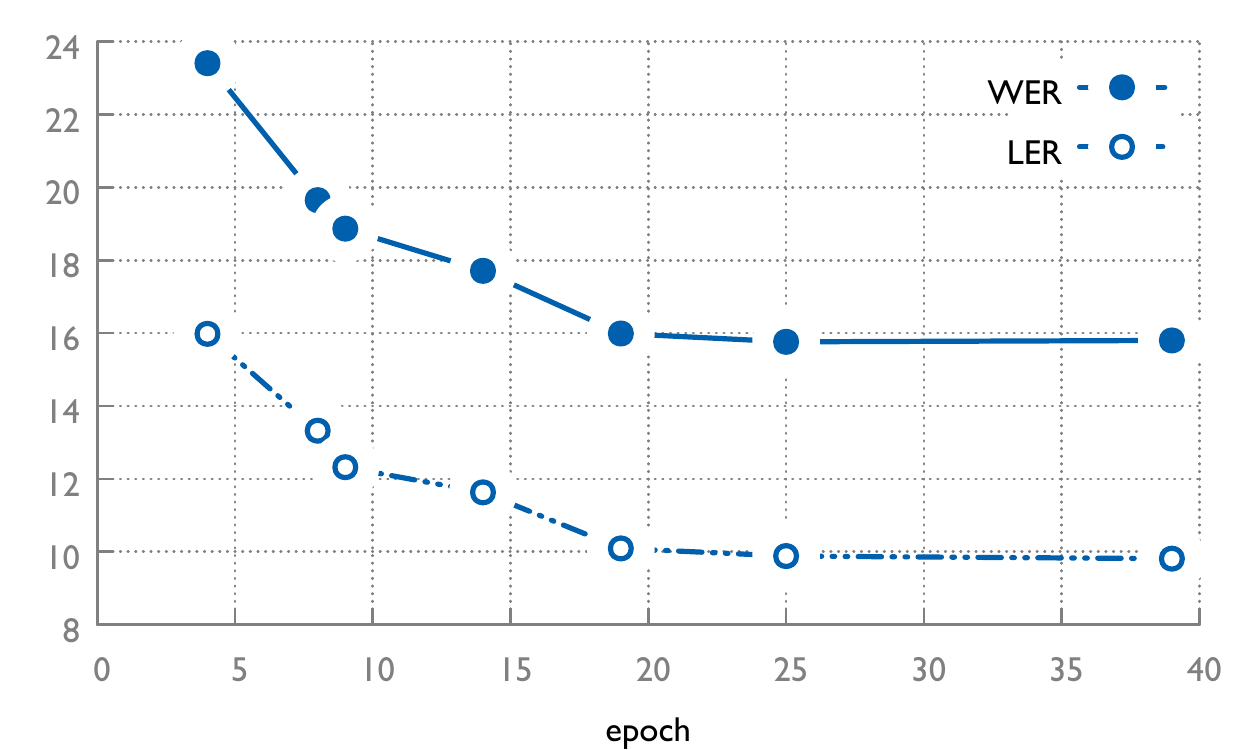}}
  \caption{
    \label{fig-ler-wer}
    LibriSpeech Letter Error Rate (LER) and Word Error Rate (WER) for the
    first training epochs of our \textsc{Low Dropout} architecture. (a) is on
    \texttt{dev-clean}, (b) on \texttt{dev-other}.  }
\end{figure}

\subsubsection{Analysis}

Figure~\ref{fig-ler-wer} shows the LER and WER on the LibriSpeech
development sets, for the first $40$ training epochs of our \textsc{Low
  Dropout} architecture. LER and WER appear surprisingly well correlated,
both on the ``clean'' and ``other'' version of the dataset.

\begin{table}
  \caption{Comparison in LER and WER of variants of our model on (a) WSJ
    and (b) LibriSpeech. LER is computed with \emph{no} decoding. Operator
    $\max$ and $\logadd$ refer to the aggregation of beam hypotheses (see
    Section~\ref{sec-decoder}).    \vspace*{0.2cm}
  }
  \centering
  \hspace*{0.25cm}
  \subfloat[][\label{tbl-variants-wer-wsj}WSJ]{
    \scriptsize
    \begin{tabular}{lcc}
      \toprule
      & \multicolumn{2}{c}{dev93} \\
      model & LER & WER \\
      \cmidrule(lr){1-1} \cmidrule(lr){2-3}
      \textsc{Low Drop.} (ASG, $\max$) & -- & 10.0 \\
      \textsc{Low Drop.} (ASG, $\logadd$) & 7.2 & 9.8 \\
      \bottomrule
    \end{tabular}
  }
  \hspace*{1cm}
  \subfloat[][\label{tbl-variants-wer}LibriSpeech]{
    \scriptsize
    \begin{tabular}{lcccc}
      \toprule
      & \multicolumn{2}{c}{dev-clean} & \multicolumn{2}{c}{dev-other} \\
      model & LER & WER & LER & WER \\
      \cmidrule(lr){1-1} \cmidrule(lr){2-3} \cmidrule(lr){4-5}
      \textsc{Low Drop.} (ASG, logadd) & 2.7 & 4.8 & 9.8 & 15.2 \\
      \textsc{High Drop.}  (CTC) & 2.6 & 4.7 & 9.5 & 14.9 \\
      \textsc{High Drop.}  (ASG, $\max$) & --  & 4.7 & -- & 14.0 \\
      \textsc{High Drop.}  (ASG, $\logadd$) & 2.3 & 4.6 & 9.0 & 13.8 \\
      \bottomrule
    \end{tabular}
  }
\end{table}

In Table~\ref{tbl-variants-wer}, we report WERs on the LibriSpeech
development sets, both for our \textsc{Low Dropout} and \textsc{High
  Dropout} architectures. Increasing dropout regularize the acoustic model
in a way which impacts significantly generalization, the effect being
stronger on noisy speech.

Table~\ref{tbl-variants-wer-wsj} and Table~\ref{tbl-variants-wer} also
report the WER for the decoder ran with the $\max(\cdot)$ operation
(instead of $\logadd(\cdot)$ for other results) used to aggregate paths in
the beam with identical (language model, lexicon) states. It appears
advantageous (as there is no code complexity increase in the decoder -- one
only needs to replace $\max(\cdot)$ by $\logadd(\cdot)$ in the code) to use
the $\logadd(\cdot)$ aggregation.

\begin{figure*}
  \centering
  \includegraphics[width=1.0\linewidth]{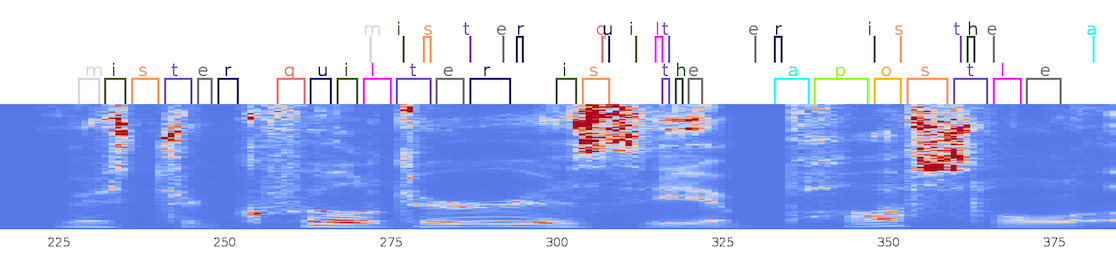}
  \caption{
    \label{fig-spectrogram}
Comparison of alignments produced by the models with CTC (top) and ASG (bottom) criterions on audio spectrogram over time (each time frame on \emph{X} axis corresponds to a 40ms window with 10 ms stride).
  }
\end{figure*}

Figure~\ref{fig-spectrogram} depicts alignments of the models with CTC and ASG criterions when forced aligned to a given target. Our analysis shows that the model with CTC criterion exhibits 500 ms delay compared to the model with ASG criterion. Similar observation was also previously noted in \cite{sak2015acoustic}.

\subsection{Comparison with other systems}

In Table~\ref{tbl-best-wer}, we compare
our system with existing phone-based and letter-based approaches on WSJ and
LibriSpeech. Phone-based acoustic state of the art models are
reported as reference. These systems output in general senones; senones are
carefully selected through a procedure involving a phonetic-context-based
decision tree built from another GMM/HMM system. Phone-based systems also
require an additional word lexicon which translates words into a sequence
of phones. Most state of the art systems also perform speaker adaptation;
iVectors compute a speaker embedding capturing both speaker and environment
information~\citep{xue2015ivectors}, while fMMLR is a two-pass decoder
technique which computes a speaker transform in the first
pass~\citep{gales1996fmllr}. Even though Table~\ref{tbl-systems} associates speaker
adaptation exclusively with phone-based systems, speaker adaptation can be also applied to letter-based systems.

State of the art performance for letter-based models on LibriSpeech is held by \textsc{Deep Speech~2}~\citep{amodei2015deep} and  \citep{zeyer2018improved} 
on noisy and clean subsets respectively. On WSJ state of the art performance is held by \textsc{Deep Speech~2}.
\textsc{Deep Speech~2} uses an acoustic model composed of a ConvNet and
a Recurrent Neural Network (RNN). \textsc{Deep Speech~2} relies on
a lot of extra speech data at training, combined with a very large 5-gram
language model at inference time to make the letter-based approach
competitive. Our system outperforms \textsc{Deep Speech~2} on clean data, 
even though our system has been trained with an order of magnitude less data.
Acoustic model in \citep{zeyer2018improved} is also based on RNNs and in addition employs attention mechanism. With LSTM language model their system shows lower WER than our, but with a simple 4-gram language model our system has slightly lower WER.

On WSJ the state of the art is a phone-based approach~\citep{chan2015rnn} which leverages an acoustic model
combining CNNs, bidirectional LSTMs, and deep fully connected neural
networks. The system also performs speaker adaptation at inference. We also
compare with existing letter-based approaches on WSJ, which are abundant in
the literature. They rely on recurrent neural networks, often
bi-directional, and in certain cases combined with ConvNet
architectures. Our system matches the best reported letter-based WSJ
performance. The Gated ConvNet appears to be very strong at modeling
complete words as it achieves $6.7\%$ WER on LibriSpeech clean data even with no decoder,
i.e. on the raw output of the neural network.

\begin{table*}
  \caption{Comparison of different near state of the art ASR systems on
    LibriSpeech. We report the type of acoustic model used for various
    systems, as well as the type of sub-word units. HMM stands for Hidden
    Markov Model, CNN for ConvNet; when not specified, CNNs are 1D. pNorm
    is a particular non-linearity~\citep{zhang2014pnorm}. We also report
    extra information (besides word transcriptions) which might be used by
    each system, including speaker adaptation, or any other domain-specific
    data. }
    \label{tbl-systems}
    \begin{center}
\setlength{\tabcolsep}{3.2pt}
    \small
    \begin{tabular}{lcccl}
      \toprule
      Paper & Acoustic Model & Sub-word & Spkr Adapt. & Extra Resources \\ \midrule
      \cite{panayotov2015librispeech} & \begin{scriptsize}HMM$+$DNN$+$pNorm\end{scriptsize} & phone & fMLLR & phone lexicon \\
      \cite{peddinti2015time} & HMM$+$CNN & phone & iVectors & phone lexicon \\
      \cite{povey2016mmi} & HMM$+$CNN & phone & iVectors & phone lexicon, \\
      &  &  &  & phone LM, data augm.\\
      \cite{ko2015audio} & \begin{scriptsize}HMM$+$CNN$+$pNorm\end{scriptsize} & phone & iVectors & phone lexicon, data augm. \\
      \midrule
        \cite{amodei2015deep} & 2D-CNN$+$RNN & letter & \emph{none} & 11.9Kh train set, \\
        &  &  &  & Common Crawl LM \\
        \cite{zhou2018policy} & CNN+GRU+policy learning & letter & \emph{none} & data augmentation \\
      \cite{zeyer2018improved} & RNN$+$attention & letter & \emph{none} & LSTM LM \\
      this paper & GLU-CNN & letter & \emph{none} & \emph{none} \\
      \bottomrule
    \end{tabular}
    \end{center}
\end{table*}

Concerning LibriSpeech, we summarize existing state of the art systems in
Table~\ref{tbl-systems}. We highlighted the acoustic model architectures,
as well as the type of underlying sub-word units.

\begin{table}
    \caption{Comparison in WER of our model with other systems on
      WSJ and LibriSpeech. Systems with $\star$ or $\dagger$ use additional data or
      data augmentation at training, respectively.}
    \label{tbl-best-wer}
    \begin{center}
      \small
    \begin{tabular}{lcccc}
      \toprule
      & WSJ eval92 & LibriSpeech test-clean & LibriSpeech test-other \\ \midrule
      \citep{povey2016mmi} & 4.3 & - & - \\
      \citep{panayotov2015librispeech} & 3.9 & 5.5 & 14.0 \\
      \citep{peddinti2015time} & - & 4.8 & - \\
      \citep{chan2015rnn} & 3.5 & - & - \\
      \citep{ko2015audio}$^{\dagger}$ & - & - & 12.5 \\
      \midrule
      \citep{hannun2014birec} & 14.1 & - & - \\
      \citep{bahdanau2016attention} & 9.3 & - & - \\
      \citep{graves2014towards} & 8.2 & - & - \\
      \citep{miao2015eesen} & 7.3 & - & - \\
      \citep{chorowski2016seqseq} & 6.7 & - & - \\
      \citep{gramctc}$^{\dagger}$ & 6.7 & - & - \\
      \citep{hori2017multi} & 5.6 & - & - \\
      \citep{zhou2018policy}$^{\dagger}$ & 5.5 & 5.7 & 15.2 \\
      \citep{amodei2015deep}$^{\star}$ & 3.6 & 5.3 & 13.3 \\
      \citep{zeyer2018improved} & - & 4.8 & 14.7 \\
      \citep{zeyer2018improved} (\emph{LSTM LM}) & - & 3.8 & 12.8 \\
      \midrule
      this paper (CTC) & - & 5.1 & 16.0 \\
      this paper (ASG) & 5.6 & 4.8 & 14.5 \\
      \bottomrule
    \end{tabular}
    \end{center}
\end{table}

\section{Conclusion}

We have introduced a simple end-to-end automatic speech recognition system,
which combines a ConvNet acoustic model with Gated Linear Units, and a
simple beam-search decoder. The acoustic model is trained to map audio
sequences to sequences of characters using a structured-output learning
approach based on a variant of CTC. Our system outperforms existing
letter-based approaches (which do not use extra data at training time
or powerful LSTM language models),
both on WSJ and LibriSpeech. 
Overall phone-based approaches are still
holding the state of the art, but our system's performance is competitive
on LibriSpeech, suggesting pronunciations is implicitly well modeled with
enough training data. Further work should include leveraging speaker
identity, training from the raw waveform, data augmentation, training with
more data, and better language models.

\bibliography{all}

\begin{thebibliography}{50}
\providecommand{\natexlab}[1]{#1}
\providecommand{\url}[1]{\texttt{#1}}
\expandafter\ifx\csname urlstyle\endcsname\relax
  \providecommand{\doi}[1]{doi: #1}\else
  \providecommand{\doi}{doi: \begingroup \urlstyle{rm}\Url}\fi

\bibitem[Amodei et~al.(2016)Amodei, Ananthanarayanan, Anubhai, Bai, Battenberg,
  Case, Casper, Catanzaro, Cheng, Chen, Chen, Chen, Chen, Chrzanowski, Coates,
  Diamos, et~al.]{amodei2015deep}
D.~Amodei, S.~Ananthanarayanan, R.~Anubhai, J.~Bai, E.~Battenberg, C.~Case,
  J.~Casper, B.~Catanzaro, Q.~Cheng, G.~Chen, J.~Chen, J.~Chen, Z.~Chen,
  M.~Chrzanowski, A.~Coates, G.~Diamos, et~al.
\newblock Deep speech 2 : End-to-end speech recognition in english and
  mandarin.
\newblock In \emph{International Conference on Machine Learning (ICML)}, pages
  173--182, 2016.

\bibitem[Bahdanau et~al.(2016)Bahdanau, Chorowski, Serdyuk, Brakel, and
  Bengio]{bahdanau2016attention}
D.~Bahdanau, J.~Chorowski, D.~Serdyuk, P.~Brakel, and Y.~Bengio.
\newblock End-to-end attention-based large vocabulary speech recognition.
\newblock In \emph{International Conference on Acoustics, Speech and Signal
  Processing (ICASSP)}, pages 4945--4949, 2016.

\bibitem[Bahl et~al.(1986)Bahl, Brown, de~Souza, and Mercer]{bahl1986icassp}
L.~R. Bahl, P.~F. Brown, P.~V. de~Souza, and R.~L. Mercer.
\newblock Maximum mutual information estimation of hidden {Markov} model
  parameters for speech recognition.
\newblock In \emph{International Conference on Acoustics, Speech and Signal
  Processing (ICASSP)}, pages 49--52, 1986.

\bibitem[Bottou(1991)]{bottou1991phd}
L.~Bottou.
\newblock \emph{Une Approche th\'eorique de l'Apprentissage Connexionniste:
  Applications \`a la Reconnaissance de la Parole}.
\newblock PhD thesis, 1991.

\bibitem[Chan and Lane(2015{\natexlab{a}})]{chan2015deep}
W.~Chan and I.~Lane.
\newblock Deep recurrent neural networks for acoustic modelling.
\newblock \emph{arXiv preprint arXiv:1504.01482}, 2015{\natexlab{a}}.

\bibitem[Chan and Lane(2015{\natexlab{b}})]{chan2015rnn}
W.~Chan and I.~Lane.
\newblock Deep recurrent neural networks for acoustic modelling.
\newblock \emph{arXiv:1504.01482}, 2015{\natexlab{b}}.

\bibitem[Chan et~al.(2016)Chan, Jaitly, Le, and
  Vinyals]{chan2016listenattendspell}
W.~Chan, N.~Jaitly, Q.~Le, and O.~Vinyals.
\newblock Listen, attend and spell: A neural network for large vocabulary
  conversational speech recognition.
\newblock In \emph{International Conference on Acoustics, Speech and Signal
  Processing (ICASSP)}, pages 4960--4964, 2016.

\bibitem[Chorowski and Jaitly(2016)]{chorowski2016seqseq}
J.~Chorowski and N.~Jaitly.
\newblock Towards better decoding and language model integration in sequence to
  sequence models.
\newblock \emph{arXiv:1612.02695}, 2016.

\bibitem[Collobert et~al.(2016)Collobert, Puhrsch, and
  Synnaeve]{collobert2016wav2letter}
R.~Collobert, C.~Puhrsch, and G.~Synnaeve.
\newblock Wav2letter: an end-to-end convnet-based speech recognition system.
\newblock \emph{arXiv:1609.03193}, 2016.

\bibitem[Dauphin et~al.(2017)Dauphin, Fan, Auli, and Grangier]{dauphin2017lm}
Y.~N. Dauphin, A.~Fan, M.~Auli, and D.~Grangier.
\newblock Language modeling with gated convolutional nets.
\newblock In \emph{International Conference on Machine Learning ({ICML})},
  2017.

\bibitem[Gales and Woodland(1996)]{gales1996fmllr}
M.~J.~F. Gales and P.~C. Woodland.
\newblock Mean and variance adaptation within the {MLLR} framework.
\newblock \emph{Computer Speech and Language}, 10\penalty0 (4):\penalty0
  249--264, 1996.

\bibitem[Gehring et~al.(2017)Gehring, Auli, Grangier, Yarats, and
  Dauphin]{gehring2017mt}
J.~Gehring, M.~Auli, D.~Grangier, D.~Yarats, and Y.~N. Dauphin.
\newblock Convolutional sequence to sequence learning.
\newblock In \emph{International Conference on Machine Learning {(ICML)}},
  2017.

\bibitem[Gibson and Hain(2006)]{gibson2006interspeech}
M.~Gibson and T.~Hain.
\newblock Hypothesis spaces for minimum {Bayes} risk training in large
  vocabulary speech recognition.
\newblock In \emph{Interspeech}, pages 2406–--2409, 2006.

\bibitem[Graves and Jaitly(2014)]{graves2014towards}
A.~Graves and N.~Jaitly.
\newblock Towards end-to-end speech recognition with recurrent neural networks.
\newblock In \emph{International Conference on Machine Learning}, pages
  1764--1772, 2014.

\bibitem[Graves et~al.(2006)Graves, Fern{\'a}ndez, Gomez, and
  Schmidhuber]{graves2006connectionist}
A.~Graves, S.~Fern{\'a}ndez, F.~Gomez, and J.~Schmidhuber.
\newblock Connectionist temporal classification: labelling unsegmented sequence
  data with recurrent neural networks.
\newblock In \emph{International Conference on Machine Learning {(ICML)}},
  pages 369--376. ACM, 2006.

\bibitem[Graves et~al.(2013)Graves, Mohamed, and Hinton]{graves2013speech}
A.~Graves, A.-R. Mohamed, and G.~Hinton.
\newblock Speech recognition with deep recurrent neural networks.
\newblock In \emph{International Conference on Acoustics, Speech and Signal
  Processing (ICASSP)}, pages 6645--6649, 2013.

\bibitem[Hannun et~al.(2014{\natexlab{a}})Hannun, Case, Casper, Catanzaro,
  Diamos, Elsen, Prenger, Satheesh, Sengupta, Coates, and Ng]{hannun2014deep}
A.~Y. Hannun, C.~Case, J.~Casper, B.~Catanzaro, G.~Diamos, E.~Elsen,
  R.~Prenger, S.~Satheesh, S.~Sengupta, A.~Coates, and A.~Y. Ng.
\newblock Deep speech: Scaling up end-to-end speech recognition.
\newblock \emph{arXiv:1412.5567}, 2014{\natexlab{a}}.

\bibitem[Hannun et~al.(2014{\natexlab{b}})Hannun, Maas, Jurafsky, and
  Ng]{hannun2014birec}
A.~Y. Hannun, A.~L. Maas, D.~Jurafsky, and A.~Y. Ng.
\newblock First-pass large vocabulary continuous speech recognition using
  bi-directional recurrent dnns.
\newblock \emph{arXiv:1408.2873}, 2014{\natexlab{b}}.

\bibitem[Heafield et~al.(2013)Heafield, Pouzyrevsky, Clark, and
  Koehn]{heafield2013scalable}
K.~Heafield, I.~Pouzyrevsky, J.~H. Clark, and P.~Koehn.
\newblock Scalable modified kneser-ney language model estimation.
\newblock In \emph{Annual Meeting of the Association for Computational
  Linguistics (ACL)}, pages 690--696, 2013.

\bibitem[Hinton et~al.(2012)Hinton, Deng, Yu, Dahl, rahman Mohamed, Jaitly,
  Senior, Vanhoucke, Nguyen, Sainath, and Kingsbury]{hinton2012deep}
G.~Hinton, L.~Deng, D.~Yu, G.~Dahl, A.~rahman Mohamed, N.~Jaitly, A.~Senior,
  V.~Vanhoucke, P.~Nguyen, T.~Sainath, and B.~Kingsbury.
\newblock Deep neural networks for acoustic modeling in speech recognition.
\newblock \emph{Signal Processing Magazine}, 29\penalty0 (6):\penalty0 82--97,
  2012.

\bibitem[Hori et~al.(2017)Hori, Watanabe, and Hershey]{hori2017multi}
T.~Hori, S.~Watanabe, and J.~R. Hershey.
\newblock Multi-level language modeling and decoding for open vocabulary
  end-to-end speech recognition.
\newblock In \emph{Automatic Speech Recognition and Understanding Workshop
  (ASRU), 2017 IEEE}. IEEE, 2017.

\bibitem[Ko et~al.(2015)Ko, Peddinti, Povey, and Khudanpur]{ko2015audio}
T.~Ko, V.~Peddinti, D.~Povey, and S.~Khudanpur.
\newblock Audio augmentation for speech recognition.
\newblock In \emph{Interspeech}, 2015.

\bibitem[LeCun et~al.(1995)LeCun, Bengio, et~al.]{lecun1995convolutional}
Y.~LeCun, Y.~Bengio, et~al.
\newblock Convolutional networks for images, speech, and time series.
\newblock \emph{The handbook of brain theory and neural networks},
  3361\penalty0 (10):\penalty0 1995, 1995.

\bibitem[Lee et~al.(1995)Lee, Ching, and Chan]{lee1995rnn}
T.~Lee, P.~Ching, and L.-W. Chan.
\newblock An rnn based speech recognition system with discriminative training.
\newblock In \emph{Fourth European Conference on Speech Communication and
  Technology}, 1995.

\bibitem[Liu et~al.(2017)Liu, Zhu, Li, and Satheesh]{gramctc}
H.~Liu, Z.~Zhu, X.~Li, and S.~Satheesh.
\newblock Gram-ctc: Automatic unit selection and target decomposition for
  sequence labelling.
\newblock \emph{CoRR}, abs/1703.00096, 2017.

\bibitem[Miao et~al.(2015)Miao, Gowayyed, and Metze]{miao2015eesen}
Y.~Miao, M.~Gowayyed, and F.~Metze.
\newblock Eesen: End-to-end speech recognition using deep {RNN} models and
  {WFST}-based decoding.
\newblock In \emph{Automatic Speech Recognition and Understanding Workshop
  (ASRU)}, 2015.

\bibitem[Mohamed et~al.(2012)Mohamed, Dahl, and Hinton]{mohamed2012acoustic}
A.-R. Mohamed, G.~E. Dahl, and G.~Hinton.
\newblock Acoustic modeling using deep belief networks.
\newblock \emph{Transactions on Audio, Speech, and Language Processing},
  20\penalty0 (1):\penalty0 14--22, 2012.

\bibitem[Panayotov et~al.(2015)Panayotov, Chen, Povey, and
  Khudanpur]{panayotov2015librispeech}
V.~Panayotov, G.~Chen, D.~Povey, and S.~Khudanpur.
\newblock Librispeech: an {ASR} corpus based on public domain audio books.
\newblock In \emph{International Conference on Acoustics, Speech and Signal
  Processing (ICASSP)}, pages 5206--5210, 2015.

\bibitem[Pascanu et~al.(2013)Pascanu, Mikolov, and Bengio]{pascanu2013rnn}
R.~Pascanu, T.~Mikolov, and Y.~Bengio.
\newblock On the difficulty of training recurrent neural networks.
\newblock In \emph{International Conference on Machine Learning {(ICML)}},
  2013.

\bibitem[Peddinti et~al.(2015{\natexlab{a}})Peddinti, Chen, Manohar, Ko, Povey,
  and Khudanpur]{peddinti2015jhu}
V.~Peddinti, G.~Chen, V.~Manohar, T.~Ko, D.~Povey, and S.~Khudanpur.
\newblock {JHU} {ASpIRE} system: Robust {LVCSR} with {TDNNs}, {iVector}
  adaptation, and {RNN}-{LMs}.
\newblock In \emph{Automatic Speech Recognition and Understanding Workshop
  (ASRU)}, 2015{\natexlab{a}}.

\bibitem[Peddinti et~al.(2015{\natexlab{b}})Peddinti, Povey, and
  Khudanpur]{peddinti2015time}
V.~Peddinti, D.~Povey, and S.~Khudanpur.
\newblock A time delay neural network architecture for efficient modeling of
  long temporal contexts.
\newblock In \emph{Interspeech}, 2015{\natexlab{b}}.

\bibitem[Povey et~al.(2016)Povey, Peddinti, Galvez, Ghahremani, Manohar, Na,
  Wang, and Khudanpur]{povey2016mmi}
D.~Povey, V.~Peddinti, D.~Galvez, P.~Ghahremani, V.~Manohar, X.~Na, Y.~Wang,
  and S.~Khudanpur.
\newblock Purely sequence-trained neural networks for {ASR} based on
  lattice-free {MMI}.
\newblock In \emph{Interspeech}, pages 2751--2755, 2016.

\bibitem[Sak et~al.(2015)Sak, de~Chaumont~Quitry, Sainath, Rao,
  et~al.]{sak2015acoustic}
H.~Sak, F.~de~Chaumont~Quitry, T.~Sainath, K.~Rao, et~al.
\newblock Acoustic modelling with cd-ctc-smbr lstm rnns.
\newblock In \emph{Automatic Speech Recognition and Understanding (ASRU), 2015
  IEEE Workshop on}, pages 604--609. IEEE, 2015.

\bibitem[Salimans and Kingma(2016)]{salimans2016wn}
T.~Salimans and D.~P. Kingma.
\newblock Weight normalization: A simple reparameterization to accelerate
  training of deep neural networks.
\newblock In \emph{Advances in Neural Information Processing Systems (NIPS)},
  pages 901--909. 2016.

\bibitem[Saon et~al.(2013)Saon, Soltau, Nahamoo, and Picheny]{saon2013speaker}
G.~Saon, H.~Soltau, D.~Nahamoo, and M.~Picheny.
\newblock Speaker adaptation of neural network acoustic models using
  {I-Vectors}.
\newblock In \emph{Automatic Speech Recognition and Understanding Workshop
  (ASRU)}, pages 55--59, 2013.

\bibitem[Saon et~al.(2015)Saon, Kuo, Rennie, and Picheny]{saon2015ibm}
G.~Saon, H.-K.~J. Kuo, S.~Rennie, and M.~Picheny.
\newblock The {IBM} 2015 english conversational telephone speech recognition
  system.
\newblock \emph{arXiv:1505.05899}, 2015.

\bibitem[Saon et~al.(2017)Saon, Kurata, Sercu, Audhkhasi, Thomas, Dimitriadis,
  Cui, Ramabhadran, Picheny, Lim, et~al.]{saon2017english}
G.~Saon, G.~Kurata, T.~Sercu, K.~Audhkhasi, S.~Thomas, D.~Dimitriadis, X.~Cui,
  B.~Ramabhadran, M.~Picheny, L.-L. Lim, et~al.
\newblock English conversational telephone speech recognition by humans and
  machines.
\newblock \emph{arXiv:1703.02136}, 2017.

\bibitem[Senior et~al.(2014)Senior, Heigold, Bacchiani, and
  Liao]{senior2014gmm}
A.~Senior, G.~Heigold, M.~Bacchiani, and H.~Liao.
\newblock {GMM}-free {DNN} training.
\newblock In \emph{International Conference on Acoustics, Speech and Signal
  Processing (ICASSP)}, pages 5639--5643, 2014.

\bibitem[Sercu et~al.(2016)Sercu, Puhrsch, Kingsbury, and LeCun]{sercu2015very}
T.~Sercu, C.~Puhrsch, B.~Kingsbury, and Y.~LeCun.
\newblock Very deep multilingual convolutional neural networks for {LVCSR}.
\newblock In \emph{International Conference on Acoustics, Speech and Signal
  Processing (ICASSP)}, pages 4955--4959, 2016.

\bibitem[Soltau et~al.(2014)Soltau, Saon, and Sainath]{soltau2014joint}
H.~Soltau, G.~Saon, and T.~N. Sainath.
\newblock Joint training of convolutional and non-convolutional neural
  networks.
\newblock In \emph{International Conference on Acoustics, Speech and Signal
  Processing (ICASSP)}, pages 5572--5576, 2014.

\bibitem[Srivastava et~al.(2014)Srivastava, Hinton, Krizhevsky, Sutskever, and
  Salakhutdinov]{srivastava2014dropout}
N.~Srivastava, G.~Hinton, A.~Krizhevsky, I.~Sutskever, and R.~Salakhutdinov.
\newblock Dropout: A simple way to prevent neural networks from overfitting.
\newblock \emph{Journal of Machine Learning Research (JMLR)}, 15\penalty0
  (Jun):\penalty0 1929--1958, 2014.

\bibitem[Steinbiss et~al.(1994)Steinbiss, Tran, and
  Ney]{steinbiss1994improvements}
V.~Steinbiss, B.-H. Tran, and H.~Ney.
\newblock Improvements in beam search.
\newblock In \emph{International Conference on Spoken Language Processing
  (ICSLP)}, volume~94, pages 2143--2146, 1994.

\bibitem[Tang et~al.(2017)Tang, Lu, Kong, Gimpel, Livescu, Dyer, A.~Smith, and
  Renals]{tang2017segmental}
H.~Tang, L.~Lu, L.~Kong, K.~Gimpel, K.~Livescu, C.~Dyer, N.~A.~Smith, and
  S.~Renals.
\newblock End-to-end neural segmental models for speech recognition.
\newblock \emph{IEEE Journal of Selected Topics in Signal Processing},
  11:\penalty0 1254--1264, 2017.

\bibitem[Waibel(1989)]{waibel1989tdnn}
A.~Waibel.
\newblock Modular construction of time-delay neural networks for speech
  recognition.
\newblock \emph{Neural Computation}, 1\penalty0 (1):\penalty0 39--46, 1989.

\bibitem[Woodland and Young(1993)]{woodland1993htk}
P.~C. Woodland and S.~J. Young.
\newblock The {HTK} tied-state continuous speech recogniser.
\newblock In \emph{Eurospeech}, 1993.

\bibitem[Xiong et~al.(2017)Xiong, Droppo, Huang, Seide, Seltzer, Stolcke, Yu,
  and Zweig]{xiong2017microsoft}
W.~Xiong, J.~Droppo, X.~Huang, F.~Seide, M.~Seltzer, A.~Stolcke, D.~Yu, and
  G.~Zweig.
\newblock The {Microsoft} 2016 conversational speech recognition system.
\newblock In \emph{International Conference on Acoustics, Speech and Signal
  Processing (ICASSP)}, pages 5255--5259, 2017.

\bibitem[Xue et~al.(2014)Xue, Abdel{-}Hamid, Jiang, Dai, and
  Liu]{xue2015ivectors}
S.~Xue, O.~Abdel{-}Hamid, H.~Jiang, L.~Dai, and Q.~Liu.
\newblock Fast adaptation of deep neural network based on discriminant codes
  for speech recognition.
\newblock \emph{Transactions on Audio, Speech and Language Processing},
  22\penalty0 (12):\penalty0 1713--1725, 2014.

\bibitem[Zeyer et~al.(2018)Zeyer, Irie, Schl{\"u}ter, and
  Ney]{zeyer2018improved}
A.~Zeyer, K.~Irie, R.~Schl{\"u}ter, and H.~Ney.
\newblock Improved training of end-to-end attention models for speech
  recognition.
\newblock \emph{arXiv:1805.03294}, 2018.

\bibitem[Zhang et~al.(2014)Zhang, Trmal, Povey, and Khudanpur]{zhang2014pnorm}
X.~Zhang, J.~Trmal, D.~Povey, and S.~Khudanpur.
\newblock Improving deep neural network acoustic models using generalized
  maxout networks.
\newblock In \emph{International Conference on Acoustics, Speech and Signal
  Processing (ICASSP)}, pages 215--219, 2014.

\bibitem[Zhou et~al.(2018)Zhou, Xiong, and Socher]{zhou2018policy}
Y.~Zhou, C.~Xiong, and R.~Socher.
\newblock Improving end-to-end speech recognition with policy learning.
\newblock In \emph{International Conference on Acoustics, Speech and Signal
  Processing (ICASSP)}, 2018.

\end{thebibliography}
\bibliographystyle{abbrvnat}

\end{document}